\documentclass[energies,article,accept,moreauthors,pdftex]{mdpi} 

\usepackage{graphicx}
\usepackage{subcaption}
\usepackage{balance}
\usepackage{url}
\usepackage{hhline}
\usepackage{rotating}
\usepackage{amssymb}
\usepackage{tabu,multirow}
\usepackage{xcolor}

\usepackage{multirow}
\usepackage{cancel}
\usepackage{amsmath}
\usepackage[normalem]{ulem}
\usepackage[ruled,vlined]{algorithm2e}

\firstpage{1} 
\pubvolume{13}
\issuenum{7}
\articlenumber{1726}
\pubyear{2020}
\copyrightyear{2020}
\history{Received: 9 March 2020; Accepted: 1 April 2020; Published: 4 April 2020}
\updates{yes} % If there is an update available, un-comment this line

\Title{Application of Genetic Algorithm for More Efficient Multi-Layer Thickness Optimization in Solar Cells}

\Author{Premkumar Vincent $^{1,\dagger}$, Gwenaelle Cunha Sergio $^{1,\dagger}$, Jaewon Jang $^{1}$, In Man Kang $^{1}$, Jaehoon~Park~$^{2}$, Hyeok Kim $^{3}$, Minho Lee $^{4}$ and Jin-Hyuk Bae $^{1,*}$} %Please carefully check the accuracy of names and affiliations. Changes will not be possible after proofreading.

\AuthorNames{Premkumar Vincent, Gwenaelle Cunha Sergio, Jaewon Jang, In Man Kang, Philippe Lang, Hyeok Kim, Jaehoon Park, Muhan Choi, Minho Lee, and Jin-Hyuk Bae}

\address{%
$^{1}$ \quad School of Electronics Engineering, Kyungpook National University, 80 Daehakro, Bukgu, Daegu~41566,~Korea; 2014600014@knu.ac.kr (P.V.); gwena.cs@gmail.com (G.C.S.); j1jang@knu.ac.kr (J.J.);
imkang@ee.knu.ac.kr (I.M.K.) \\%country names has been changed according to mdpi country list
$^{2}$ \quad College of Software, Hallym University, Chuncheon 24252, Korea; jaypark@hallym.ac.kr\\
$^{3}$ \quad Department of Electrical and Computer Engineering, University of Seoul, 163 Seoulsiripdaero, Dongdaemun-gu, Seoul 02504, Korea; hyeok.kim@uos.ac.kr\\
$^{4}$ \quad Department of Artificial Intelligence, Kyungpook National University, 80 Daehakro, Bukgu, Daegu~41566,~Korea; mholee@knu.ac.kr}

\corres{Correspondence: jhbae@ee.knu.ac.kr}

\firstnote{Co-first authors with equal contribution.}
\abstract{Thin-film solar cells are predominately designed similar to a stacked structure. Optimizing the layer thicknesses in this stack structure is crucial to extract the best efficiency of the solar cell. The~commonplace method used in optimization simulations, such as for optimizing the optical spacer layers' thicknesses, is the parameter sweep. Our simulation study shows that the implementation of a meta-heuristic method like the genetic algorithm results in a significantly faster and accurate search method when compared to the brute-force parameter sweep method in both single and multi-layer optimization. While other sweep methods can also outperform the brute-force method, they do not consistently exhibit $100\%$ accuracy in the optimized results like our genetic algorithm. We have used a well-studied P3HT-based structure to test our algorithm. Our best-case scenario was observed to use $60.84\%$ fewer simulations than the brute-force method.}%it is not allowed to have any link cited in abstract part, please remove it


\begin{document}
%%%%%%%%%%%%%%%%%%%%%%%%%%%%%%%%%%%%%%%%%%

\section{Introduction}
\label{S:intro}
Simulations of optoelectronic devices have helped to understand and design better optimized structures with efficiencies nearing the theoretical maximum. Lucio {et. al.} analyzed the possibility of achieving the limits of c-Si solar cells through such simulations~\cite{doi:10.1080/23746149.2018.1548305}. Simulations have reduced the time it takes for researchers to find optimized device structure. However, the~most common way to obtain results over a large range of a parameter's values is through parameter sweep method. This brute-force method is ineffective in most cases where the user only requires the end optimized device structure. Genetic algorithm (GA) is an optimization algorithm in artificial intelligence based on Darwin's evolution and natural selection theory, in~which the fittest outcome survives~\cite{darwin2004origin,man1996genetic}. This algorithm sets an environment with a random population and a function, which is called the fitness function, that scores each individual of that population. The~environment then selects individuals to become the parents of the next generation through a selection process. The~next generation of individuals (children of the previous generation's parents) is obtained via a crossover method. Similar to natural mutation in genes of the offspring, the~new generation's individuals can also suffer mutation in their genes. After~several generations, the~population converges to the individuals representing the optimal solution. Through the application of genetic algorithm, Jafar-Zanjani {et. al.} designed a binary-pattern reflect-array for highly efficient beam steering~\cite{Jafar-Zanjani2018} and Tsai {et. al.} was able to beam-shape the laser to obtain up to $90\%$ uniformity in intensity distribution~\cite{Tsai:15}. Genetic algorithm has been used by Donald {et. al.} to improve the focusing of light propagating through a scattering medium~\cite{Conkey:12} and by \mbox{Wen {et. al.}} for designing highly coherent optical fiber in the mid-infrared spectral range~\cite{Zhang:09}. It~has also been used for designing nanostructures to improve light absorption in solar cells. \mbox{Chen {et. al.}} were able to surpass the Yablonovitch Limit using genetic algorithms to design light trapping nanostructures~\cite{Wang2013}. Rogério {et. al.} used a genetic algorithm to design surface structures on a Si solar cell to increase the short-circuit current density obtained from it~\cite{doi:10.1063/1.5078745}.

In this article, we have demonstrated the optimization of an organic solar cell through the optimization of the optical spacer layers. Traditionally, finite difference time domain (FDTD) method was used to simulate the ideal short-circuit current density ($J_{sc}$) of the solar cell through the Lumerical, FDTD solutions software similar to our previous reported study~\cite{KIM2020152202}. Parameter sweep or brute-force method was used to vary the thickness of the optical spacer layers of the solar cell. At~the optimized layer thicknesses, the~solar cell will be observed to have the highest $J_{sc}$ output. Although~not computationally intensive for a single-layer optimization, the~number of simulations expands as in Equation \eqref{eq:brute_force_complexity} for multi-layer optimization problems.
\begin{equation} \label{eq:brute_force_complexity}
    N = n_1 \cdot  n_2 \cdot n_3 ...NN
\end{equation}
where $N$ is the total number of simulations and $n_1$, $n_2$, and~$n_3$ are the number of simulations performed for layers 1, 2, and~3, and~NN is the total number of layers (NN > 3), respectively. %computation~time

%The algorithm is similar to that of Darwin's natural selection hypothesis in which the strongest shall survive. 
% Because of its inspiration in the evolution theory, GA can also be referred as an evolutionary algorithm. 
To alleviate the brute-force method's limitations, we propose the use of GA. This article then aims to heuristically assert the hypothesis that GA is a more efficient approach than brute-force algorithms in tasks such as optimizing optoelectronic device~structures.

\section{Methodology} \label{S:method}
\begin{figure}[H]
    \centering
    \includegraphics[width=0.4\textwidth]{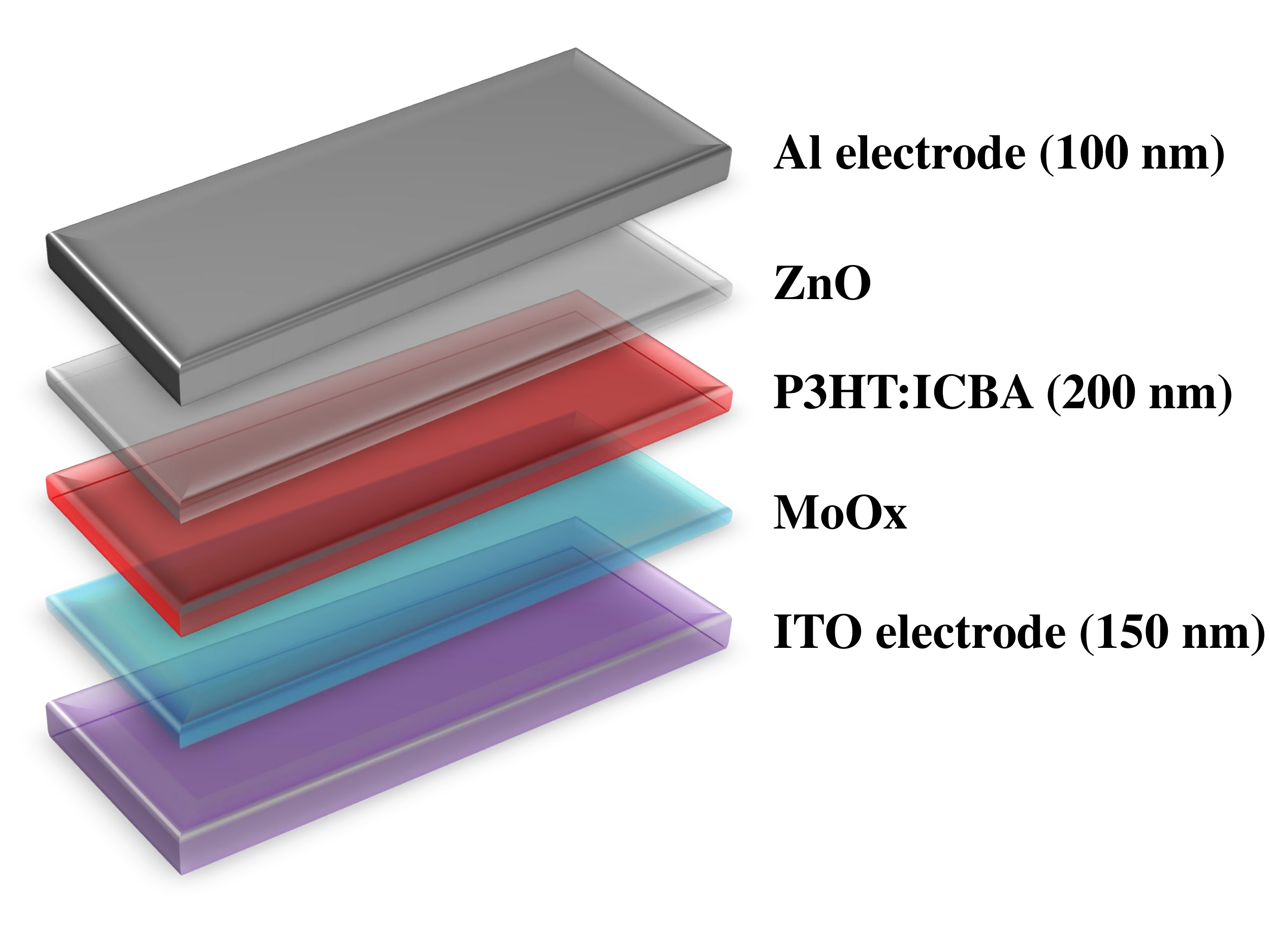}
    \caption{Solar cell device structure. The~electron transport layer is ZnO, while the hole transport layer is MoOx. Both also act as optical spacer layers as they affect the distribution of light inside the device. We optimized these optical spacer layers for maximizing photon absorption inside the active layer of the solar~cell.}
    \label{fig:figure 1}
\end{figure}

Figure~\ref{fig:figure 1} shows the device structure that was
constructed in Lumerical, FDTD solutions. It~consists of a 150 nm indium tin oxide (ITO) similar to our previous study~\cite{cite3_vincent2017dependence}. The~Al thickness was set to 100 nm as light gets completely reflected from the Al electrode at this thickness. Any~further increase in its thickness would not affect the outcome of our optical simulation. The~active layer, poly (3-hexylthiophene) (P3HT): indene-$C_{60}$ bisadduct (ICBA), was designed to be 200 nm as it exhibited good efficiency in our previous study~\cite{cite1_vincent2018indoor}. The~charge transport layers, zinc oxide (ZnO) and Molybdenum oxide (MoOx), also act as optical spacer layers and are variable quantities in our simulation. In~order to qualify as an optical spacer, the~layer material should have the refractive index properties to shift the light induced electric field inside the solar cell structure by varying its thickness. ZnO was already shown to be a good optical spacer in our previous research~\cite{cite3_vincent2017dependence}. While a ITO/PEDOT:PSS/P3HT:PCBM/ZnO/Al is a more commonly used solar cell structure~\cite{C0JM02354J}, PEDOT:PSS does not have the suitable refractive index to control the distribution of electric field for the wavelengths under consideration. MoOx was found to be a suitable hole transport layer substitute to the PEDOT:PSS by a previously reported study by Bohao {et. al.}~\cite{doi:10.1021/ph4000168}. MoOx was also found to have good optical spacer properties in our simulations and so it was used as the other optical spacer~layer.

{The simulation was done in perpendicular illumination onto the solar cell as done for device simulations. This method does not simulate the real outdoor operation of the solar cell, and~thus, the~optimized thickness values calculated in this article would not be applicable in the view of energy yield optimization. However, our study was a comparison of the algorithms, and~our conclusion should be consistent.}The solar cell layers were stacked along the \emph{y}-axis in the FDTD software. The~incident light was set up as a plane wave with the spectral intensity of AM1.5G. The~plane wave's propagation direction was the same axis along which the solar cell's layers were stacked. It was incident through the ITO electrode. Periodic boundaries were set along the \emph{x}-axis and perfectly matched layers were used along the \emph{y}-axis to set the FDTD boundary conditions. The~device was meshed good enough for the results to converge. Further simulation setup details are published elsewhere~\cite{cite1_vincent2018indoor,cite3_vincent2017dependence}. To~simulate the ideal $J_{sc}$, $100\%$ internal quantum efficiency was~assumed. 

\unskip

\subsection{Brute Force}
\begin{figure}[H]
    \centering
    \includegraphics[width=1\textwidth]{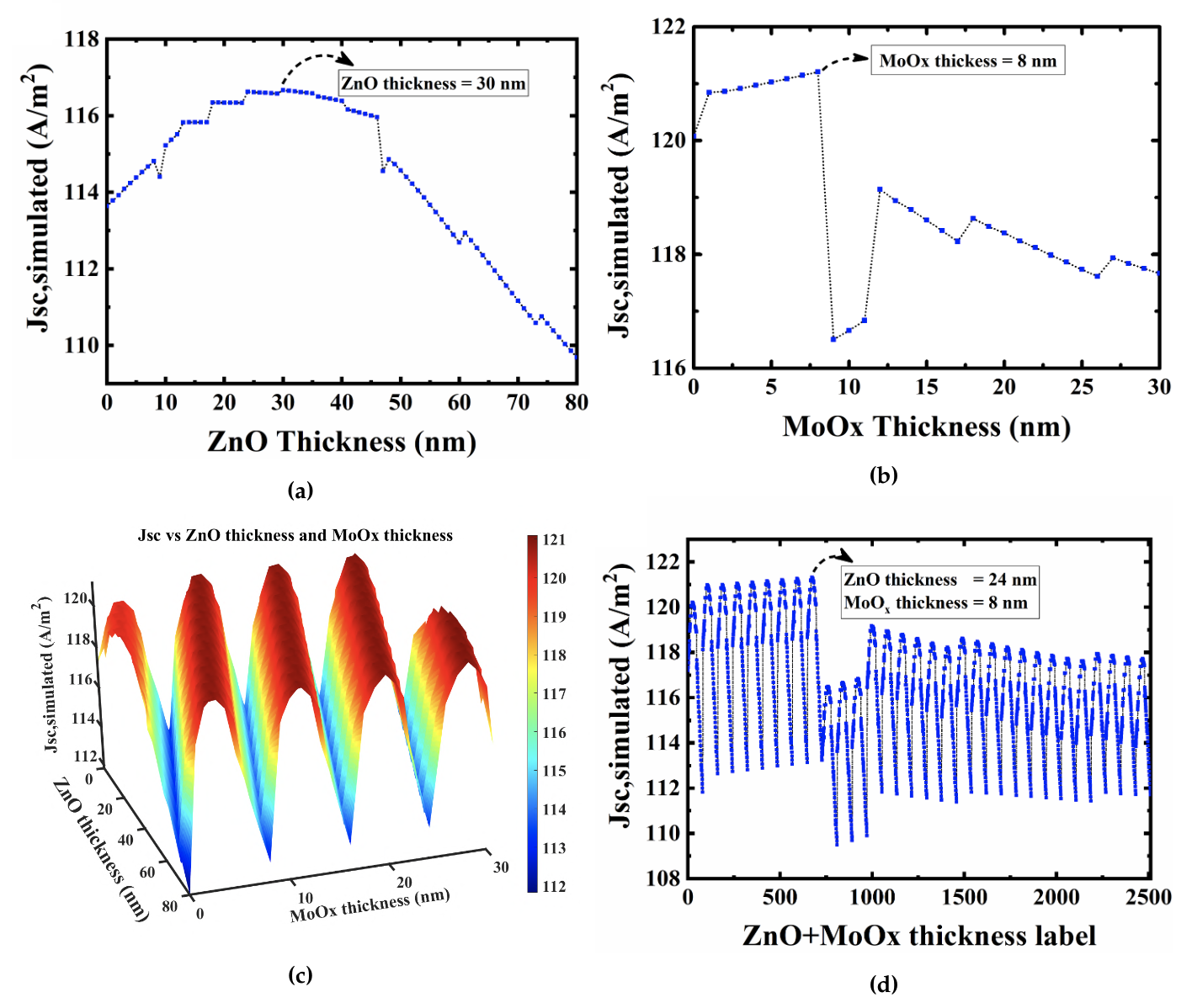}
    \caption{\textls[-25]{Brute-force method results: (\textbf{a}) single ZnO layer optimization, (\textbf{b}) single MoOx layer optimization, (\textbf{c}) multiple ZnO and MoOx layers optimization, (\textbf{d}) 2D data representation of (\textbf{c}) using labels pointing to the ZnO and MoOx layer thickness combinations for the ease of~computation.}}%please unify the format in the main text and in the figure.
\label{fig:figure 2}
\end{figure}

Using the software's parameter sweep option, we simulated our device structure according to three sections. The~first section optimized only the ZnO layer, while keeping the MoOx layer at 10 nm thickness. The~second simulation section consists of optimizing only the MoOx layer, while the ZnO layer was fixed at 30 nm. Our final section optimized both optical spacer layers together. The~results of the brute-force method are provided in Figure~\ref{fig:figure 2}.

Using parameter sweep to simulate every possible combination is time-consuming. Figure~\ref{fig:figure 2} contains multiple local maxima and minima points. This makes it tedious to extrapolate the optimal thickness from fewer simulation points. Due to this, every possible simulation point is required to find the optical result. This way of obtaining the optimal result is termed as brute-force method henceforth. In~order to make a 2-layer optimization problem similar to that of the single-layer optimization, we replaced the optical spacers' thickness combinations with a label in Figure~\ref{fig:figure 2}d, effectively converting 3D data to a 2D data. The~label number is given by \(label\: number = max(ZnO\: thickness) \times MoOx\: thickness + MoOx\: thickness + 1\).%please confirm whether the 

%\begin{figure}[htbp!]
    %\begin{subfigure}{0.5\textwidth}
    %\includegraphics[width=1\linewidth]{fig2a_ZnO_brute_force} 
    %\caption{\label{fig:figure 2a}}
    %\end{subfigure}
    %\begin{subfigure}{0.5\textwidth}
    %\includegraphics[width=1\linewidth]{fig2b_MoOx_brute_force}
    %\caption{\label{fig:figure 2b}}
    %\end{subfigure}
    %~
    %\begin{subfigure}{0.5\textwidth}
    %\includegraphics[width=1\linewidth]{fig2c_ZnO_MoO3} 
    %\caption{\label{fig:figure 2c}}
    %\end{subfigure}
    %\begin{subfigure}{0.5\textwidth}
    %\includegraphics[width=1\linewidth]{fig2d_ZnO_MoOx_2D_using_label}
    %\caption{\label{fig:figure 2d}}
    %\end{subfigure}
    %\includegraphics[width=0.5\textwidth]{fig2_ZnO_brute_force} 
%\caption{Brute-force method results: (a) single ZnO layer optimization, (b) single MoOx layer optimization, (c) multiple ZnO and MoOx layers optimization, (d) 2D data representation of (c) using labels pointing to the ZnO and MoOx layer thickness combinations for the ease of computation.}
%\label{fig:figure 2}
%\end{figure}

\subsection{Genetic~Algorithm}
% The fittest individuals according to the score given by that function will persevere.
To alleviate the computational and time inefficiencies of the brute-force method, we found inspiration in Darwin's natural selection theory and proposed the use of a genetic algorithm~\cite{man1996genetic} for our optimization problem. Consider that there is a random population and their adaptability to the environment is given by a fitness function. In~our optimization problem, the~fitness function was to maximize $J_{sc}$ output from the FDTD simulation. Each population contains several individual chromosomes, which in turn consists of an array of bits called genes. Different selection methods are used to choose certain chromosomes in each generation (see Section~\ref{s:selection}) in order to reproduce off-springs using a crossover method (see Section~\ref{s:crossover}). To~further mimic biology, there is also a probability that a chromosome might suffer mutation, which is provided by the mutation rate, which allows the algorithm to escape local minima in the data. In~the end, the~fittest members of the population prevail, meaning that the algorithm converges to the optimal~solution.

A step-by-step of the works of the genetic algorithm is shown in Algorithm \ref{alg:genetic_algorithm}. The~initial population $pop$ of size $p$ is randomly selected from the search space, which in our case is the maximum and minimum thickness of the optical spacer layers. The~first iteration of the algorithm then starts. For~each population value, the~GA calls the FDTD software to simulate and extract the $J_{sc}$ result. The~fitness function ($J_{sc}$) is applied to all individuals in the population and ranked from the highest $J_{sc}$ to the lowest. A~selection method is then applied taking into consideration each chromosome and their respective fitness score. After~selecting the parents responsible for the next generation, they reproduce to obtain the next generation, a step detailed in Algorithm \ref{alg:reproduction}. Please note that in our algorithm, the~fittest individual in the current generation is cloned to be part of the next generation. The~current generation is then updated with the next generation, with~the algorithm continuing until the maximum number of generations has been~reached.

\vspace{6pt}
\begin{algorithm}[H]
 \KwData{$max\_generation$, $population\_size\: (p)$, $mutation\_prob$, $search\_space$ (range of population)}
 \KwResult{Best individual and fitness value (optimal solution)}
 $pop \gets$ random initial population of size $p$ from $search\_space$\;
 \While{$generation <= max\_generation$}{
   $fitness\_score$ $\gets$ Fitness($pop$)\;
   $fitness\_score\_sorted$, $pop\_sorted$ $\gets$ Rank($fitness\_score$, $pop$)\;
   $new\_pop \gets$ \textbf{Cloning} of the $fittest$\;
   $next\_parents \gets$ Selection($fitness\_score\_sorted$, $pop\_sorted$)\;
   $new\_pop \gets$ [$new\_pop$, Reproduction($p-1$, $next\_parents$, $mutation\_prob$)]\;
   $pop \gets new\_pop$\;
   $generation$ += 1\;
 }
 Save $pop\_sorted$ and $fitness\_score\_sorted$ in $.mat$\;
 \caption{Genetic~Algorithm}\label{alg:genetic_algorithm}
\end{algorithm}
\vspace{6pt}
%\vspace{0.3cm}
%\vspace{0.3cm}
%\begin{algorithm}[ht!]
% \KwData{$max\_gen$, $p$, $mutation\_prob$, $search\_space$ (range of population)}
% \KwResult{Best individual and fitness value (optimal solution)}
% $pop \gets$ random initial population of size $p$ from $search\_space$\;
% \While{$gen <= max\_gen$}{
%   $f$ $\gets$ Fitness($pop$)\;
%   $f\_sorted$, $pop\_sorted$ $\gets$ Rank($f$, $pop$)\;
%   $next\_parents \gets$ Selection($f\_sorted$, $pop\_sorted$)\;
%   $new\_pop \gets$ Reproduction($p-1$, $next\_parents$, $mutation\_prob$)\;
%   $new\_pop \gets$ [$new\_pop$, Best($f\_sorted$)]\;
%   $pop \gets new\_pop$\;
%   $gen$ += 1\;
% }
% Save $pop\_sorted$ and $f\_sorted$ in $.mat$\;
% \caption{Genetic Algorithm - IS THIS BETTER FOR THE READER?}\label{alg:genetic_algorithm}
%\end{algorithm}
%\vspace{0.3cm}

As for the reproduction algorithm, detailed in Algorithm \ref{alg:reproduction}, the~first step is to allocate an empty array for the new population. The~second step is to select two parents to produce $C$ children. Due to the crossover method used and mutation ratio, the~same parents can reproduce different children. The~new population is then updated with the new children obtained through crossover and~mutation.%please confirm whether the algorithm is correct?

\vspace{6pt}
\begin{algorithm}[H]
 \KwData{$next\_parents$, $p$, $mutation\_prob$}
 \KwResult{New population}
 $new\_pop \gets []$\;
 \For{i=1 to $p$ / 2}{
      $parent1 \gets$ $next\_parents$(i)\;
      $parent2 \gets$ $next\_parents$(len($next\_parents$) - i + 1)\;
      Create $C$ children from $parent1$ and $parent2$ with Crossover and Mutations\;
      $new\_pop \gets$ $new\_pop$ + new children
 }
 \caption{Reproduction~Algorithm}\label{alg:reproduction}
\end{algorithm}
\vspace{0.3cm}

\subsubsection{Selection~Methods}
\label{s:selection}
Four selection methods were examined in this work: random, tournament, roulette wheel, and~breeder. For~better clarity of the following explanations, the~fittest individual means the individual with highest fitness score, since our problem is a maximization problem.
%the fittest individual means the individual with lowest fitness score, in a minimization problem, or the one with highest fitness score, in a maximization problem.
%Suppose the algorithm is minimizing the fitness function, then the individual with lowest fitness wins the round. This individual will become a parent for the next generation.

\subsubsection*{Random} This is the simplest selection method since it does not incorporate selection criteria. This method consists of randomly selecting individuals to be the next generation's parents, with~no regards to the fitness function. Because~of this Monte-Carlo-like approach, it can take very long for the algorithm to~converge.

\subsubsection*{Tournament} Tournament Selection~\cite{fang2010review} samples $k$ individuals with replacement from a population of $p$ and applies the fitness function to those individuals in order to select the one with best fitness score, also known as the fittest individual. One can think of this method as a battle of the fittest, where $k$ individuals face each other in tournament fashion to decide the fittest. The~fittest individuals from each tournament round will then constitute the parents responsible for forming the next~generation.

%To choose the fittest among the $N$ individuals, the fitness function is applied to them. Suppose the algorithm is minimizing the fitness function, then the individual with lowest fitness wins the round. This individual will become a parent for the next generation.

% Source: https://www.tutorialspoint.com/genetic_algorithms/genetic_algorithms_parent_selection.htm
\subsubsection*{Roulette Wheel} Roulette-Wheel Selection (RWS)~\cite{jebari2013selection} is a popular way of parent selection in which individuals have a fitness-proportionate probability of being selected. In~that way, if~an individual is very fit, it has a higher chance of being chosen, otherwise their chance is lower. 
%Figure~\ref{fig:roulette} offers a different approach to explaining this method. All individuals are organized in a chart with probability between 0 and 1, proportionate to their respective fitness values, with sum equals to 1. A random number, also between 0 and 1, is then generated, and wherever it falls on the roulette, that individual will be chosen. For example, in Figure~\ref{fig:roulette}, if the random number generated is 0.3, individual 3 will be chosen. 
%A form of fitness-proportionate selection in which the chance of an individual's being selected is proportional to the amount by which its fitness is greater or less than its competitors' fitness
%``Fitness Proportionate Selection is one of the most popular ways of parent selection. In this every individual can become a parent with a probability which is proportional to its fitness. Therefore, fitter individuals have a higher chance of mating and propagating their features to the next generation. Therefore, such a selection strategy applies a selection pressure to the more fit individuals in the population, evolving better individuals over time."

%\begin{figure}[htbp!]
%\centering
%  \subfloat{\includegraphics[width=.5\textwidth]{img_rws_table.pdf}} %\vspace{-0.4cm}
%  %\hfill
%  \hspace{0.5cm}
% \subfloat{\includegraphics[width=.3\textwidth]{img_rws_chart_blue.pdf}}
%  \caption{Roulette Wheel Selection (RWS) for parent selection. Individuals 1 and 6 have higher chance to be selected given that their fitness score is the highest, whereas individual 2 has the lowest.}
%  \label{fig:roulette}
%\end{figure}

\subsubsection*{Breeder}  Breeder Selection~\cite{muhlenbein1993predictive} follows the same strategy used for breeding animals and plants, where the goal is to preserve certain desired properties from the parents in their children. This is achieved by conserving the genetic material from the fittest individuals while still giving some mutation leeway by adding a few random individuals (lucky few) to the mix of parents for the next~generation.

% Check: https://en.wikipedia.org/wiki/Crossover_(genetic_algorithm)#Uniform_crossover
\subsubsection{Crossover}
\label{s:crossover}
Crossover is the reproduction method in genetic algorithms, and~it consists of choosing the parts of each parent that will be present in their~child.

\subsubsection*{Uniform}
\begin{figure}[H]
    \centering
    \includegraphics[width=0.7\textwidth]{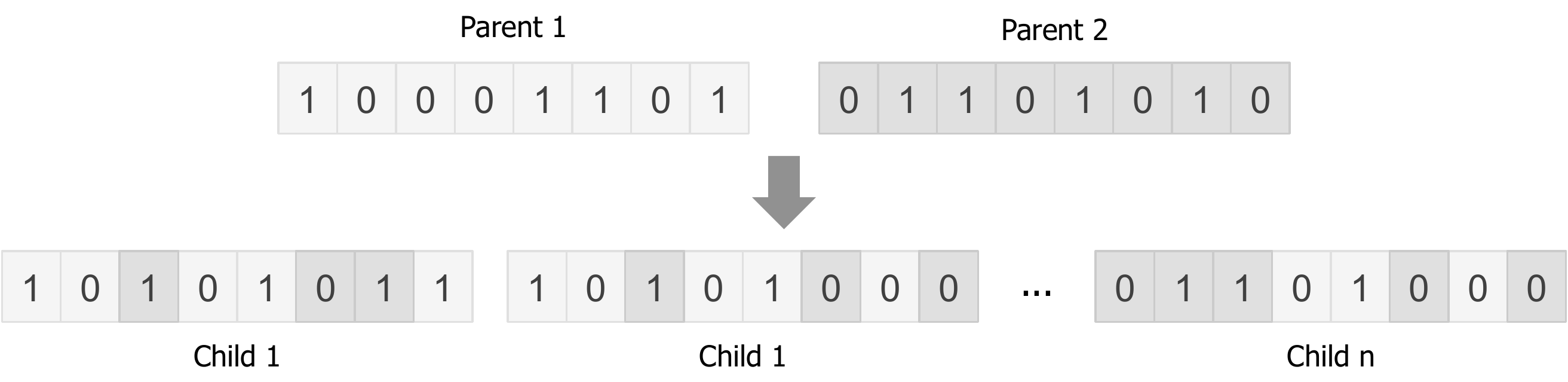}
    \caption{Uniform crossover: bitwise~reproduction.}
    \label{fig:crossover}
\end{figure}

A uniform crossover is shown in Figure~\ref{fig:crossover}. In~this type of crossover, each bit is chosen from one of the parents with probability of 0.5. The~advantage of this method is that the same parents can form many children with a more diverse set of~genes.

% Other options are one-point, two-point and $k$-point.
%\paragraph{One-point, two-point, and $k$-point} As the name indicates, in one-point crossover, each parent is divided in one point and the first half is taken from parent 1 and the second from parent 2. Two-point has two points of division and k-point $k$. One and two-point methods are simpler, but they result in less diverse off-springs.

\subsubsection*{{$k$}-Point}
\begin{figure}[H]
    \centering
    \includegraphics[width=0.5\textwidth]{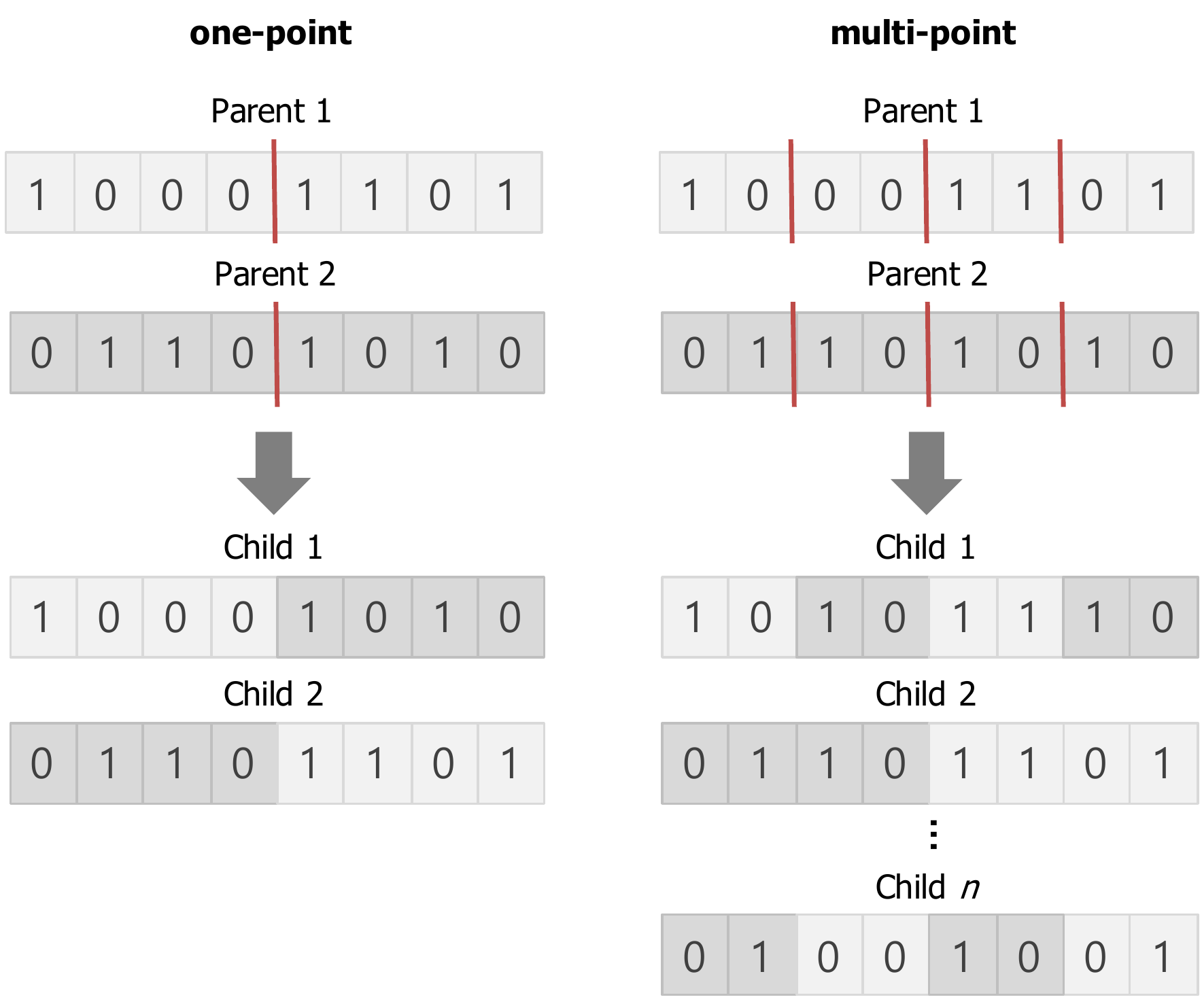}
    \caption{$k$-point crossover: blockwise~reproduction.}
    \label{fig:crossover_k_point}
\end{figure}

In a $k$-point crossover, each parent is divided into $k$ segments. These segments are then chosen with equal probability to compose the new chromosomes for their children. This crossover, shown in Figure~\ref{fig:crossover_k_point}, can be a one-point crossover ($k=1$), or~a multi-point crossover ($1<k<N_C$, where $N_C$ is the length of a chromosome). The~disadvantage of the one-point crossover is that it is only able to generate two distinct~children.

\unskip

\subsubsection{Mutation}
\label{s:mutation}
\begin{figure}[H]
    \centering
    \includegraphics[width=0.5\textwidth]{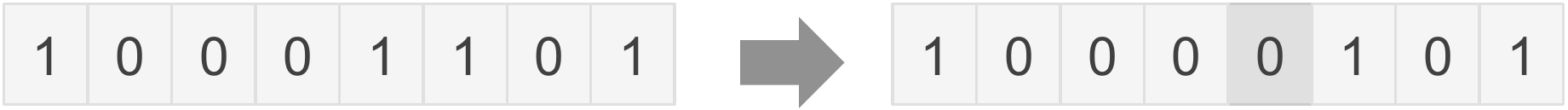}
    \caption{Mutation of a bit in a~chromosome.}
    \label{fig:mutation}
\end{figure}

This genetic operator is used to ensure genetic diversity within a group of individuals and to ensure the algorithm does not converge to a local minimum. The~mutation operator works by flipping bits in a chromosome according to a mutation probability, as~shown in Figure~\ref{fig:mutation}. %This probability needs to be kept small so that the algorithm does not turn into a random search~method.

\section{Complexity~Analysis}
In this section, we aim to further explain our algorithm's suitability with respect to its complexity. In~brute force, the~total simulation count increases with respect to the solar cell's layers which are thickness-optimized, as~shown previously in Equation \eqref{eq:brute_force_complexity}. In~Big-O notation, which represents the upper bound for time complexity in an algorithm, the~brute-force method has algorithmic complexity as in Equation \eqref{eq:brute_force_bigO}:
\begin{equation} \label{eq:brute_force_bigO}
    O_{bruteforce} = O(n^l)
\end{equation}
where $l$ is the number of layers and $n$ is the number of fitness function evaluations for a layer. In~other words, the~complexity increases exponentially with the number of layers, and~that can be very expensive as that number~grows.

\begin{figure}[H]
    \centering
    \includegraphics[width=0.7\textwidth]{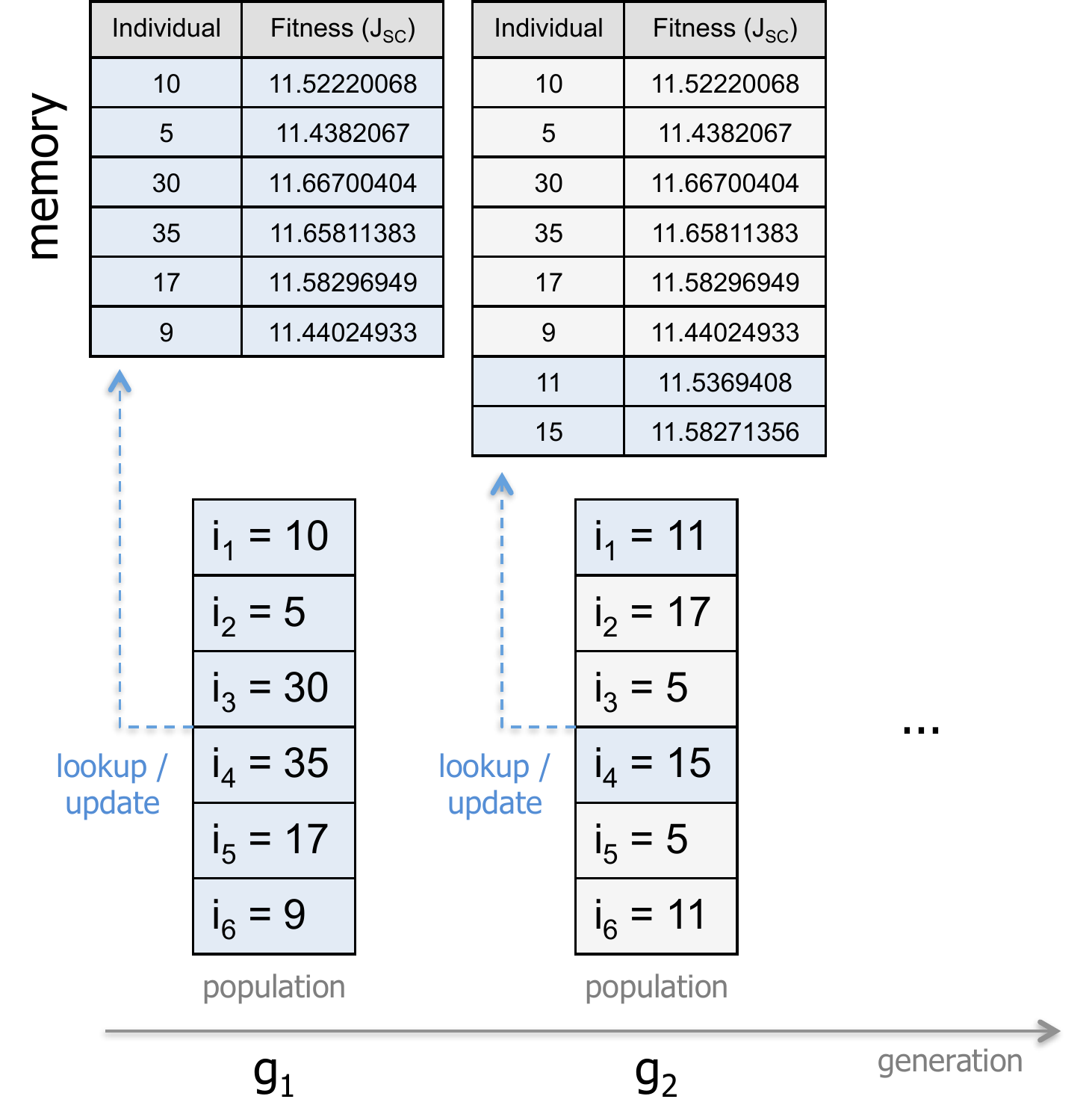}
    \caption{Genetic algorithm with dynamic programming, where blue represents values that need to be added to the lookup table in order to update it and grey represents values that have already been evaluated and need only to be copied when~necessary.}
\label{fig:ga_dynamic_programming}
\end{figure}

Our goal is to efficiently optimize layer thickness in devices composed of a single layer and multiple layers alike. Hence, we used genetic algorithm for our global optimization requirement. GA can converge to an optimal solution by evaluating less individuals than the brute-force method. However, in~the classical approach, the~same individual might be evaluated multiple times. To~prevent this redundancy from happening, we implement GA with dynamic programming. In~this approach, there is a dynamic dictionary, also called lookup table, which serves as memory to store the already evaluated individuals and their respective fitness scores. The~dictionary is said to be dynamic because it may change size if the algorithm receives an individual whose fitness score has not been calculated yet. This approach is illustrated with an example in Figure~\ref{fig:ga_dynamic_programming}, where in every generation, or~iteration, the~algorithm searches the memory for a desired individual. If~this individual has already been computed, its fitness score can simply be used by the algorithm. Otherwise, the~fitness function is evaluated, and the lookup table is~updated.
 
\section{Results and~Discussion}
\label{S:results}
Since GA is a stochastic algorithm, we calculated the average number of simulations required by the GA and its standard deviation from 5000 repeated runs per section. The~accuracy of the data discussed below are all $100\%$, which means that all the 5000 runs converged to the optimal solution. We~have discussed three sections below: single layer---ZnO thickness optimization; single layer---MoOx thickness optimization; and~multiple layers---concurrent ZnO and MoOx thickness~optimization. 

\subsection{Single~Layer}
\unskip
\subsubsection{ZnO Optical Spacer~Layer}
For the single ZnO optical spacer layer optimization, we fixed the MoOx layer thickness as 10 nm. We applied the brute force and the genetic algorithm to our task and solved it within the same thickness limits of 0 to 80 nm. The~best $J_{sc}$, which is also the fitness value, was obtained when the ZnO thickness was optimized to 30 nm (as shown in Figure~\ref{fig:figure 2}a). We have compared the total number of simulations and their respective accuracies for different selection methods. The~initial population size, generations count, and~mutation probability were the factors that were iterated in order to find the conditions that would use the least number of simulations to optimize the device structure. The~population size was varied from 10 to 80 in increments of 10, the~generation count from 10 to 100 in increments of 10, and~the mutation probability from 5 to 100 in increments of 5. While the brute-force method required 81 simulations in total, the~number of simulations required by the genetic algorithm was dependent on the selection method and initialization parameters used. The~initialization parameters which provided the least average number of simulations from the 5000 simulations, while keeping the accuracy at $100\%$, is provided in Table~\ref{tbl:ZnOtable}.

\begin{table}[H]
\caption{Comparison results for ZnO single-layer~optimization.}
\label{tbl:ZnOtable}
\centering
\scalebox{1.0}{
\begin{tabular}{ l  c c c c }
\toprule
& \multicolumn{4}{c} {\textbf{Brute-Force Method: Number of Simulations = }\textbf{81}} \\
& \multicolumn{4}{c} {\textbf{Optimized ZnO Thickness = 30 nm}} \\
& \multicolumn{4}{c}{\textbf{Selection Method}} \\ \midrule
\textbf{Parameter} & \textbf{Random} & \textbf{Roulette} & \textbf{Tournament} & \textbf{Breeder} \\ \midrule
Population & 20 & 80 & 70 & 60 \\
Generation & 40 & 10 & 30 & 10 \\
Mutation prob (\%) & 80 & 15 & 60 & 75 \\ \midrule
\textbf{Mean (simulations)} & $78.42 \pm\: 1.82$ & $80.47 \pm\: 0.50$ & \textbf{78.16 $\pm\: 1.65$} & $79.17 \pm\: 1.37$ \\ %does it need to be bold? please confirm
\bottomrule
\end{tabular}}
%\end{center}
\end{table}

\begin{figure}[H]
 \centering
    \includegraphics[width=0.8\textwidth]{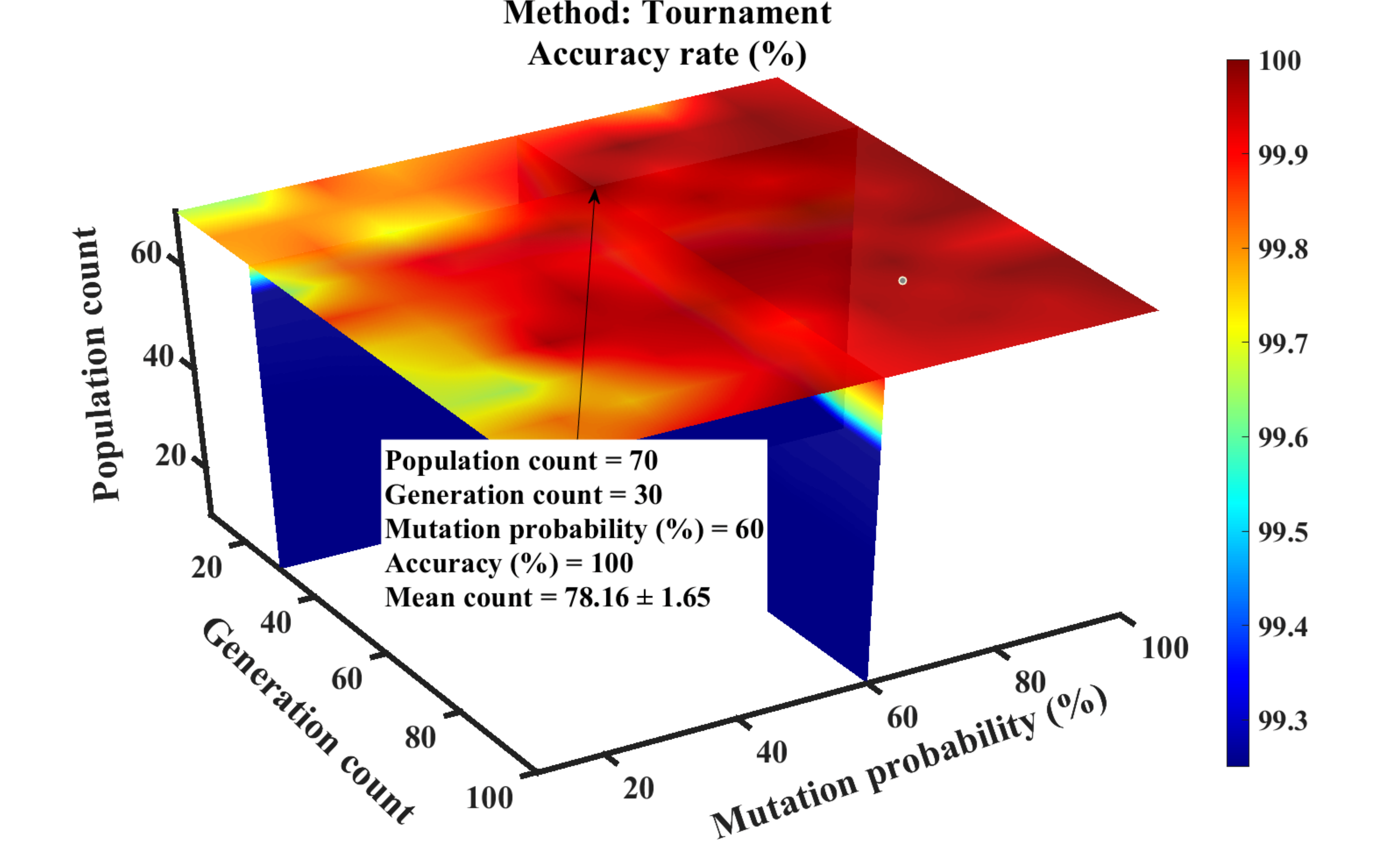}
\caption{Single ZnO layer optimization using GA with tournament selection: accuracy ($\%$) distribution data sliced at population size of 70, generation count of 30, and~mutation probability of $60\%$.}
\label{fig:figure 6}
\end{figure}

Figure~\ref{fig:figure 6} presents the accuracy distribution over different initialization parameters. It was observed that the best result was obtained while using tournament selection model with the population size of 70, generation count of 30, and~mutation probability of $60\%$. It required $78.16 \pm\: 1.65$ simulations to reach the optimal solution. Although~the GA algorithm was observed to produce only a reduction of $2.26 \pm 2.04\%$ in the number of simulations required, this was mainly due to two optimal result points in the data. Since the device with a 24 nm ZnO layer thickness exhibited a $J_{sc}$ of 116.62 A/m$^{2}$  and the one with the optimal 30 nm ZnO layer thickness exhibited a near same $J_{sc}$ of 116.67 A/m$^{2}$, the~algorithm took longer to converge at the optimal structure. In~a practical scenario, however, if~both the above-mentioned structures were regarded as optimal, the~algorithm would converge with high accuracy with much lesser number of~simulations.

\unskip

\subsubsection{MoOx Optical Spacer~Layer}
For the MoOx optical spacer single-layer optimization, the~ZnO thickness was fixed at 30 nm. The~brute-force method required 31 simulations to determine the optimized layer thickness of 8 nm (Figure~\ref{fig:figure 2}b). For~finding the best initialization parameter combination, we varied the population size from 5 to 20 in increments of 5, the~generation count from 10 to 100 in increments of 10, and~the mutation probability from 5 to 100 in increments of 5. The~best case results for each selection model is provided in Table~\ref{tbl:MoOxtable}.

\begin{table}[H]
\caption{Comparison results for MoOx single-layer~optimization.}
\label{tbl:MoOxtable}
\centering
%	\begin{center}
\scalebox{1.0}{
\begin{tabular}{ l  c c c c }
\toprule
& \multicolumn{4}{c} {\textbf{Brute-Force Method: Number of Simulations =} \textbf{31}} \\
& \multicolumn{4}{c} {\textbf{Optimized MoOx Thickness = 8 nm}} \\
& \multicolumn{4}{c}{\textbf{Selection Method}} \\ \midrule
\textbf{Parameter} & \textbf{Random} & \textbf{Roulette} & \textbf{Tournament} & \textbf{Breeder} \\ \midrule
Population & 15 & 5 & 15 & 15 \\
Generation & 20 & 100 & 30 & 80 \\
Mutation prob (\%) & 80 & 75 & 75 & 80 \\ \midrule
\textbf{Mean (simulations)} & $30.91 \pm\: 0.31$ & \textbf{13.05 $\pm\: 3.24$} & $30.97 \pm\: 0.16$ & $30.97 \pm\: 0.18$ \\ 
\bottomrule
\end{tabular}}
%\end{center}
\end{table}

\begin{figure}[H]
 \centering
    \includegraphics[width=0.8\textwidth]{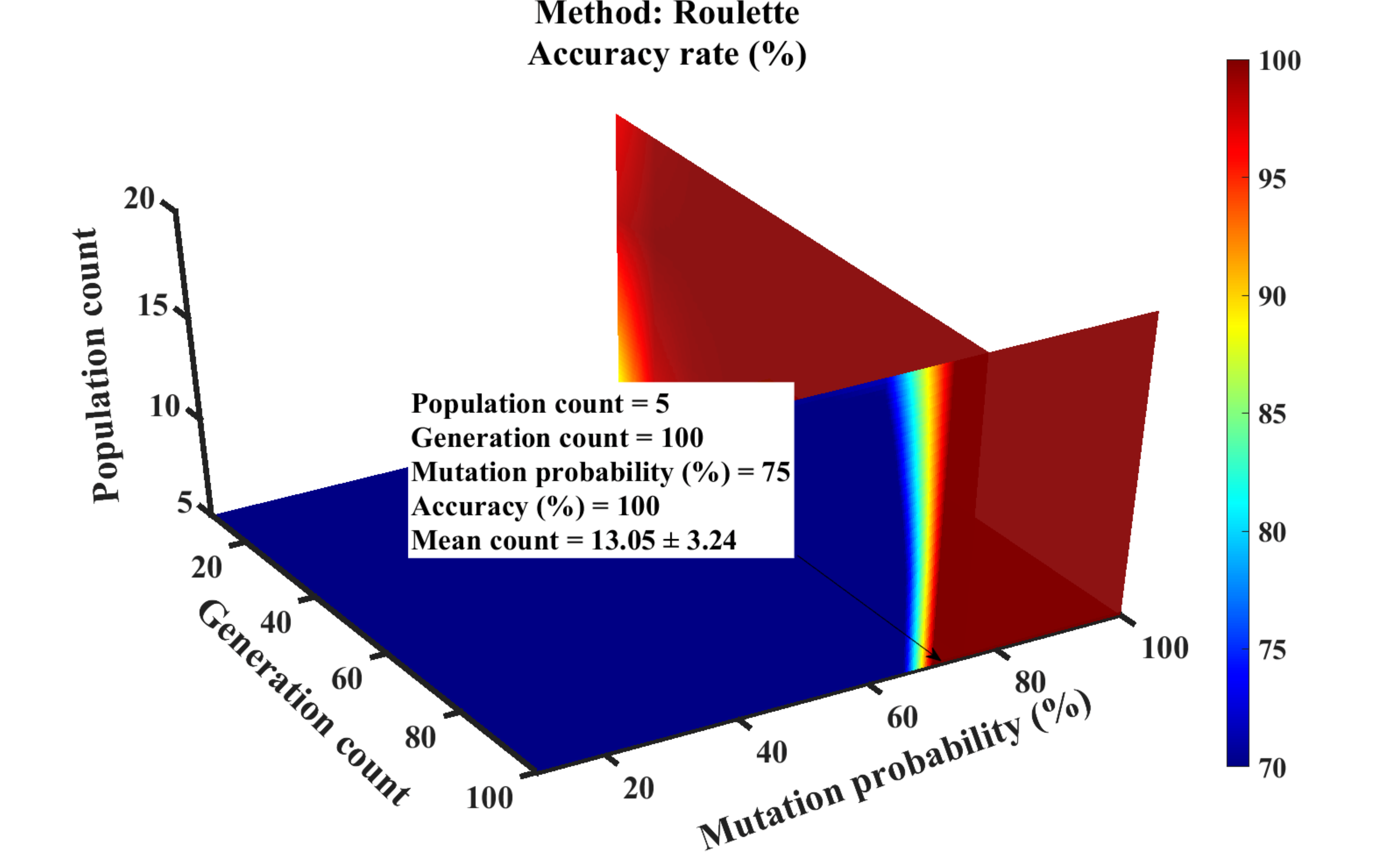}
\caption{Single MoOx layer optimization using GA with roulette selection: accuracy ($\%$) distribution data sliced at population size of 5, generation count of 100, and~mutation probability of $75\%$.}
\label{fig:figure 7}
\end{figure}

It was observed that while the roulette method was able to use $57.9 \pm 10.45\%$ fewer simulations to determine the optimal MoOx thickness, the~other methods required nearly the same number of simulations as the brute-force method. We hypothesize that the roulette method's preferential weighing of the fittest model aided in the convergence to the optimal solution faster. Figure~\ref{fig:figure 7} presents the accuracy distribution for the optimum initializing parameter~values.

%We also exhibit the advantage of evolutionary algorithm over brute-force method for multiple layer thickness optimization.

%Please note that we choose a small population $p$ in the experiments, since setting the population number as the same number being analyzed in the brute-force method nullifies the advantages of using GA. That is due to the fact that a high population means the same search space is being analyzed over and over again, and our goal is to minimize the number of times the fitness function is evaluated. Furthermore, to obtain the best combination of parameters, we perform multiple simulations with different number of population, number of generations, mutation probability, number of bits, and selection and crossover methods.

\subsection{Multi-Layer: ZnO + MoOx}
As mentioned earlier,  multi-layer optimization can take a large number of simulations. The~computational and time cost required to run these simulations are expensive. GA can be used to refine the optimization process to take as less simulations as required. The~ZnO layer thickness was incremented by 1 nm from 0 to 80 nm, while the same thickness increment was done from 0 to 30 nm for the MoOx layer. The~brute-force method used 2511 simulations to find the optimized optical spacer layer thicknesses of 24 nm ZnO and 8 nm MoOx. We converted the 3D Figure~\ref{fig:figure 2}c to a 2D Figure~\ref{fig:figure 2}d by applying labels for each ZnO and MoOx thickness combination. Thus, we have 2511 labels and the GA algorithm was applied to determine the label pointing to the optimal result. The~population size that was varied from 500 to 1500 in increments of 500, the~generation count from 10 to 100 in increments of 10, and~the mutation probability from 10 to 100 in increments of 10. Table~\ref{tbl:multilayertable} presents the result from multi-layer~optimization.

\begin{table}[H]
\caption{Comparison results for multi-layer~optimization.}
\label{tbl:multilayertable}
\centering
%\begin{center}
\scalebox{1.0}{
\begin{tabular}{ l  c c c c }
\toprule
& \multicolumn{4}{c} {\textbf{Brute-Force Method: Number of Simulations =} \textbf{2511}} \\
& \multicolumn{4}{c} {\textbf{Optimized ZnO Thickness = 24 nm, Optimized MoOx Thickness = 8 nm}} \\
& \multicolumn{4}{c}{\textbf{Selection Method}} \\ \midrule
\textbf{Parameter} & \textbf{Random} & \textbf{Roulette} & \textbf{Tournament} & \textbf{Breeder} \\ \midrule
Population & 500 & 1000 & 500 & 500 \\
Generation & 90 & 90 & 80 & 90 \\
Mutation prob (\%) & 10 & 90 & 20 & 50 \\ \midrule
\textbf{Mean (simulations)} & $2391.34 \pm\: 38.13$ & \textbf{1758.77 $\pm\: 39.75$} & $2428.84 \pm\: 34.57$ & $2256.80 \pm\: 70.15$ \\ 
\bottomrule
\end{tabular}}
%\end{center}
\end{table}

\begin{figure}[H]
 \centering
    \includegraphics[width=0.8\textwidth]{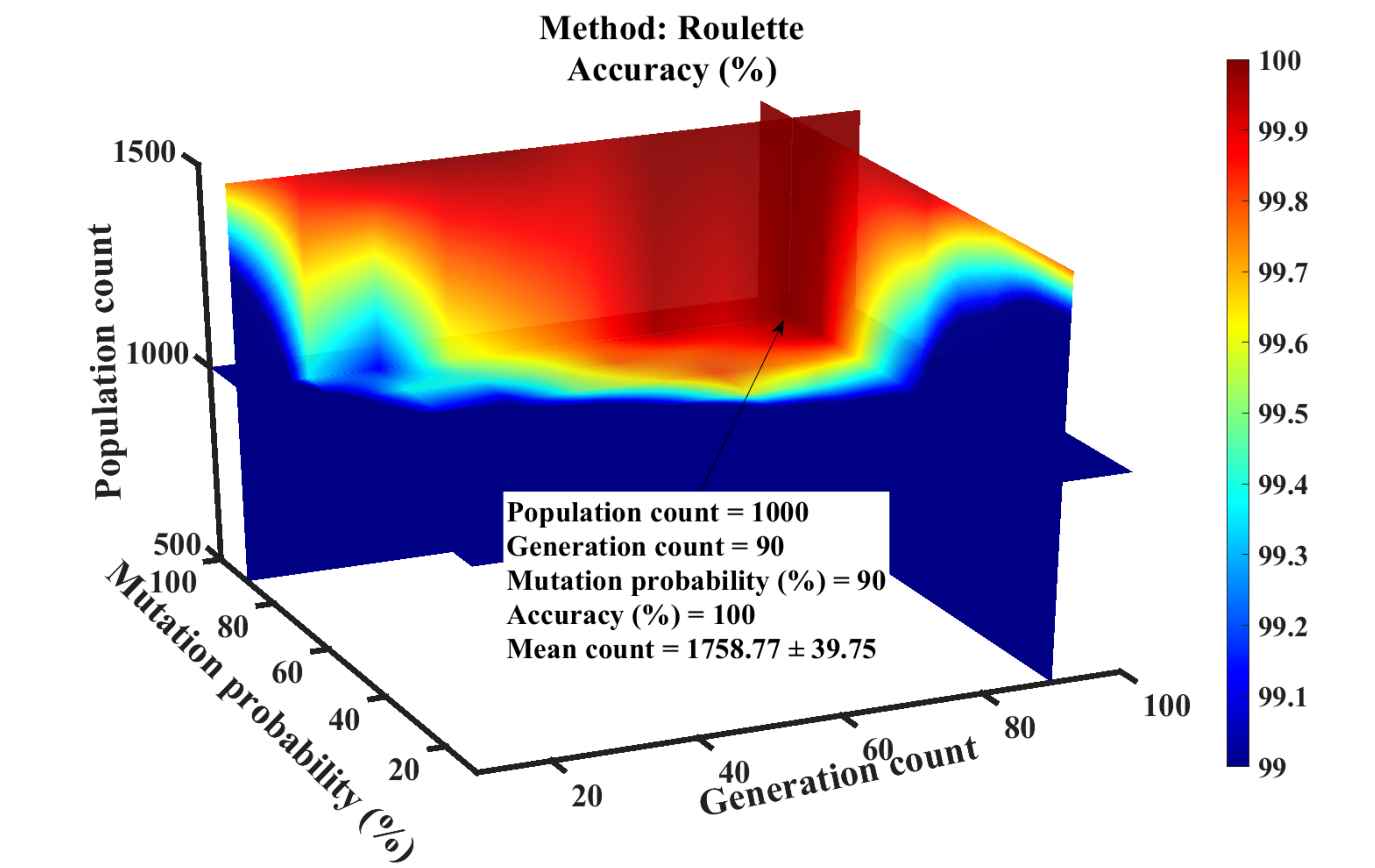}
\caption{Multi-layer optimization using GA with roulette selection: accuracy ($\%$) distribution data sliced at population size of 1000, generation count of 90, and~mutation probability of $90\%$.}
\label{fig:figure 8}
\end{figure}

Figure~\ref{fig:figure 8} presents a $29.96 \pm\: 1.58\%$ reduction in the number of simulations required by the roulette selection method to obtain the optimal solution. As~the complexity became higher, having more randomness in the population through a high mutation probability rate aided constructively to reduce the number of simulations required. Due to this, the~roulette selection method was able to show a best average number of simulation count of $1758.77 \pm\: 39.75$ from 5000 repetitive~runs.

\unskip

%%%%%%%%%%%%%%%%%%%%%%%%%%%%%%%%%%%%%%%%%%

\subsection{Performance Comparison: Uniform vs $K$-Point Crossover~Methods}
The roulette method demonstrates satisfactory performance in the ZnO layer thickness optimization problem and the best performances for both MoOx and multi-layer thickness optimization problems. These results are obtained using the uniform crossover genetic operator, whereas performances with the $k$-point crossover method remain uncharted. This section aims to fill that gap and provide a performance comparison between uniform and $k$-point crossover methods. We use $k$ values of 1, 2, and~4 for our $k$-point crossover~method.

Table~\ref{tbl:crossover_comparison} shows the average number of simulations required to solve the optimization problem when using uniform versus when using $k$-point crossover. It also compares the result obtained through the various crossover methods to the chromosome's binary bit size. In~the MoOx case with $k=1$ and $bits=12$, the~chromosome is divided into segments of 6 bits. Since the maximum number of bits required to define the MoOx layer thicknesses is 5 bits, one of the parents does not pass on its chromosome to the child. Due to this, there is no genetic variation between the parents and children, essentially meaning that the children are copies of one parent, hence the solution did not converge. We also observed that in most cases using k-point crossover, the~best results were obtained when the chromosome was segmented into segments of 4 bits in length. The~exception to this were the MoOx case with 5-bit chromosomes and the ZnO case with 12-bit chromosomes. Uniform crossover and 4-point crossover proved better in these cases, respectively. We hypothesize that the poorer performance at other $k$-point values was linked to the amount of variation in the chromosome. Lower $k$-point values meant fewer segmentations in the parent chromosome, which in turn meant fewer genetic variations occurred, leading to a slower convergence of the result. However, at the same time, higher $k$-point value meant high genetic variation which also lead to slower~convergence.

\begin{table}[H]
\caption{Comparison results for uniform and $k$-point crossover~methods.}
\label{tbl:crossover_comparison}
\centering
%\begin{center}
\scalebox{1.0}{
\begin{tabular}{ c  c c c c  c }
\toprule
& \multicolumn{4}{c}{\textbf{Average Number of Simulations}} &  \\
& \multicolumn{4}{c}{\textbf{Crossover Methods}} &  \\ \midrule
\textbf{Layers} & \textbf{Uniform} & \textbf{1-point} & \textbf{2-point} & \textbf{4-point} & \textbf{Bits} \\ \midrule 
MoOx & \textbf{13.05 $\pm$ 3.24} & 17.83 $\pm$ 2.54 & 20.89 $\pm$ 2.84 & 14.73 $\pm$ 3.06 & 5\\ % 11.41\% decay
MoOx & \textbf{13.05 $\pm$ 3.24}  & 13.83 $\pm$ 1.69 & 19.88 $\pm$ 3.74 & 20.99 $\pm$ 3.72 & 8 \\
MoOx & 13.05 $\pm$ 3.24 & - * & \textbf{12.14 $\pm$ 1.26} & 19.05 $\pm$ 2.13 & 12 \\ 
ZnO & 80.47 $\pm$ 0.50 & \textbf{70.76 $\pm$ 0.84} & 75.72 $\pm$ 1.44 & 74.00 $\pm$ 1.96 & 8 \\ % 12.07\% improvement
ZnO & 80.47 $\pm$ 0.50 & 74.04 $\pm$ 1.58 & 76.37 $\pm$ 1.56 & \textbf{73.90 $\pm$ 1.51} & 12 \\
ZnO-MoOx & 1758.77 $\pm$ 39.75	& 1701.78 $\pm$ 15.34 & \textbf{1140.06 $\pm$ 37.11} & 1355.23 $\pm$ 68.35 & 12 \\ % 35.18\% improvement
\bottomrule
\end{tabular}}
\begin{tabular}{ccc}
\multicolumn{1}{c}{\footnotesize * Solution did not converge.}
\end{tabular}

%\\
%\hspace{-10cm} \footnotesize{*Solution did not converge}
%%\end{center}
\end{table}

The best initialization parameters for the optimization problems discussed in this article are tabulated in Table~\ref{tbl:best_init_params}.

\begin{table}[H]
\caption{Initialization parameters for best~performances.} \label{tbl:best_init_params}
\centering
%\begin{center}
\scalebox{1.0}{
\begin{tabular}{ c  c c c  c c c }
\toprule
 & \textbf{Average Number} & \textbf{Crossover} &  & \multicolumn{3}{c}{\textbf{Initialization Parameters}} \\
\textbf{Layers} & \textbf{of Simulations} & \textbf{Method} & \textbf{Bits} & \textbf{Population} & \textbf{Generation Count} & \textbf{Mutation Rate} \\ \midrule
MoOx & 12.14 $\pm$ 1.26 & 2-point & 12 & 10 & 30 & 10 \\
ZnO & 70.76 $\pm$ 0.84 & 1-point & 8 & 70 & 10 & 15 \\
MoOx + ZnO & 1140.06 $\pm$ 37.11 & 2-point & 12  & 500 & 70 & 70 \\ \bottomrule
\end{tabular}}
%\end{center}
\end{table}

%The sweet spot was found to be k=2 for 12-bit chromosomes. We checked our hypothesis by using 8-bit chromosomes for ZnO and MoOx single-layer optimization problems. Multi-layer optimization problem required a minimum of 12-bits and so it was not tested with 8-bits. We found that the both ZnO and MoOx optimization problems were able to achieve their best results when the parent chromosomes were segmented as sets of 4-bits; i.e., k=1 when chromosome bit size is 8, and k=2 when the chromosome bit size is 12.}

% since it is hard to withhold important information and patterns if each bit varies differently

%we say that the more bits required to represent a parent, the more k-point shows improvement?

%So MoOx uniform an k=4 being similar is good for us, because it makes sense since k=5 is uniform, so we can just say that the reason is that it is becoming a uniform crossover

%%%%%%%%%%%%%%%%%%%%%%%%%%%%%%%%%%%%%%%%%%

\section{Conclusions}
\label{S:conclusion}
We have demonstrated that the genetic algorithm can perform better than the conventional parameter sweep used in simulations. In~our best-case scenario, it exhibited no loss in accuracy, while outperforming the brute-force method by up to $57.9\%$ with the correct initialization parameters. In~the worst-case scenario, the~GA utilized the same number of simulations as the brute-force method, demonstrating that it cannot be outperformed by brute force. We found that the best selection method was the roulette-wheel selection. For~uniform crossover method, it exhibited a satisfactory performance for ZnO layer optimization and an outstanding performance for the MoOx and 2-layer optimization problems. When using k-point crossover, the~roulette method was able to further decrease the number of simulations required to converge to the optimal result. In~conclusion, we~were able to reduce the average number of simulations required for MoOx layer optimization to 12.14 (brute force = 31), ZnO~layer optimization to 70.76 (brute force = 81), and~ZnO-MoOx layers optimization to 1140.06 (brute force = 2511). The~GA is dependent on its initialization parameters and the selection method chosen. This article does not discuss an automated way to assign these parameters as it is not in its research scope. However, the~results suggest that there is possibility for greatly refining the parameter sweep method using genetic algorithms as shown with both single and multi-layer optimization of the solar cell~structure. Code and additional results are at \url{https://github.com/gcunhase/GeneticAlgorithm-SolarCells}.

%%%%%%%%%%%%%%%%%%%%%%%%%%%%%%%%%%%%%%%%%%
%\section{Patents}
%This section is not mandatory, but may be added if there are patents resulting from the work reported in this manuscript.
\newpage
%%%%%%%%%%%%%%%%%%%%%%%%%%%%%%%%%%%%%%%%%%
\vspace{6pt} 

%%%%%%%%%%%%%%%%%%%%%%%%%%%%%%%%%%%%%%%%%%
%% optional
%\supplementary{The following are available online at \linksupplementary{s1}, Figure~S1: title, Table S1: title, Video S1: title.}

% \supplementary{The following are available at \linksupplementary{s1}, Figure~S1: title, Table S1: title, Video S1: title. A supporting video article is available at doi: link.}

%%%%%%%%%%%%%%%%%%%%%%%%%%%%%%%%%%%%%%%%%%
% For research articles with several authors, a short paragraph specifying their individual contributions must be provided. The following statements should be used ``conceptualization, X.X. and Y.Y.; methodology, X.X.; software, X.X.; validation, X.X., Y.Y. and Z.Z.; formal analysis, X.X.; investigation, X.X.; resources, X.X.; data curation, X.X.; writing--original draft preparation, X.X.; writing--review and editing, X.X.; visualization, X.X.; supervision, X.X.; project administration, X.X.; funding acquisition, Y.Y.'', please turn to the  \href{http://img.mdpi.org/data/contributor-role-instruction.pdf}{CRediT taxonomy} for the term explanation. Authorship must be limited to those who have contributed substantially to the work reported.
\authorcontributions{Conceptualization, P.V.; Methodology, P.V. and G.C.S.; Software, P.V., and~G.C.S.; Validation, J.-H.B., J.J., I.M.K., J.P., H.K., and~M.L.; Formal analysis, P.V., G.C.S., J.-H.B., J.J., H.K., and~J.P.; Investigation, P.V., and~G.C.S.; Resources, H.K., and~J.-H.B.; Data curation, P.V., and~G.C.S.; Writing---original draft preparation, P.V., and~G.C.S.; Writing---review and editing, P.V., G.C.S., J.-H.B.; Visualization, P.V., and~G.C.S.; Supervision, J.-H.B., and~M.L.; Project administration, J.-H.B.; Funding acquisition, J.-H.B. All authors have read and agreed to the published version of the manuscript.}

%%%%%%%%%%%%%%%%%%%%%%%%%%%%%%%%%%%%%%%%%%
% Please add: ``This research received no external funding'' or ``This research was funded by NAME OF FUNDER grant number XXX.'' and  ``The APC was funded by XXX''. Check carefully that the details given are accurate and use the standard spelling of funding agency names at \url{https://search.crossref.org/funding}, any errors may affect your future funding.
%\funding{}

%%%%%%%%%%%%%%%%%%%%%%%%%%%%%%%%%%%%%%%%%%
% In this section, you can acknowledge any support given which is not covered by the author contribution or funding sections. This may include administrative and technical support, or donations in kind (e.g., materials used for experiments).
\funding{This research was supported by Basic Science Research Program through the National Research Foundation of Korea (NRF) funded by the Ministry of Science and ICT (2018R1A2B6008815), and~also by the BK21 Plus project funded by the Ministry of Education, Korea (21A20131600011).}%changed to fundig, please confirm.

%%%%%%%%%%%%%%%%%%%%%%%%%%%%%%%%%%%%%%%%%%
\conflictsofinterest{The authors declare no conflict of interest.%Declare conflicts of interest or state ``The authors declare no conflict of interest.'' The authors must identify and declare any personal circumstances or interest that may be perceived as inappropriately influencing the representation or interpretation of reported research results. Any role of the funders in the design of the study; in the collection, analyses or interpretation of data; in the writing of the manuscript, or~in the decision to publish the results must be declared in this section. If~there is no role, please state ``The funders had no role in the design of the study; in the collection, analyses, or~interpretation of data; in the writing of the manuscript, or~in the decision to publish the results''.
} 

%%%%%%%%%%%%%%%%%%%%%%%%%%%%%%%%%%%%%%%%%%
%% optional
\abbreviations{The following abbreviations are used in this manuscript:\\

\noindent 
\begin{tabular}{@{}ll}
GA & Genetic Algorithm\\
FDTD & Finite Difference Time Domain\\
ZnO & Zinc Oxide\\
MoOx & Molybdenum Oxide\\
RWS & Roulette-Wheel Selection\\
\end{tabular}}

%%%%%%%%%%%%%%%%%%%%%%%%%%%%%%%%%%%%%%%%%%
%% optional
\appendixtitles{no} %Leave argument "no" if all appendix headings stay EMPTY (then no dot is printed after "Appendix A"). If~the appendix sections contain a heading then change the argument to "yes".
% \appendix
% \section{}
% \unskip
% \subsection{}
% The appendix is an optional section that can contain detail and data supplemental to the main text. For example, explanations of experimental details that would disrupt the flow of the main text, but nonetheless remain crucial to understanding and reproducing the research shown; figures of replicates for experiments of which representative data is shown in the main text can be added here if brief, or as Supplementary data. Mathematical proofs of results not central to the paper can be added as an appendix.

% \section{}
% All appendix sections must be cited in the main text. In the appendixes, Figures, Tables, etc. should be labeled starting with `A', e.g.,~Figure~A1, Figure~A2, etc. 


%=====================================
% References, variant A: internal bibliography
%=====================================
\reftitle{References}
% \begin{thebibliography}{999}
% % Reference 1
% \bibitem[Author1(year)]{ref-journal}
% Author1, T. The title of the cited article. {\em Journal Abbreviation} {\bf 2008}, {\em 10}, 142--149.
% % Reference 2
% \bibitem[Author2(year)]{ref-book}
% Author2, L. The title of the cited contribution. In {\em The Book Title}; Editor1, F., Editor2, A., Eds.; Publishing House: City, Country, 2007; pp. 32--58.
% \end{thebibliography}

% The following MDPI journals use author-date citation: Arts, Econometrics, Economies, Genealogy, Humanities, IJFS, JRFM, Laws, Religions, Risks, Social Sciences. For those journals, please follow the formatting guidelines on http://www.mdpi.com/authors/references
% To cite two works by the same author: \citeauthor{ref-journal-1a} (\citeyear{ref-journal-1a}, \citeyear{ref-journal-1b}). This produces: Whittaker (1967, 1975)
% To cite two works by the same author with specific pages: \citeauthor{ref-journal-3a} (\citeyear{ref-journal-3a}, p. 328; \citeyear{ref-journal-3b}, p.475). This produces: Wong (1999, p. 328; 2000, p. 475)

%=====================================
% References, variant B: external bibliography
%=====================================
%\externalbibliography{yes}
%\bibliography{sample}

%%%%%%%%%%%%%%%%%%%%%%%%%%%%%%%%%%%%%%%%%%
%% optional
%\sampleavailability{Samples of the compounds ...... are available from the authors.}

%% for journal Sci
%\reviewreports{\\
%Reviewer 1 comments and authors’ response\\
%Reviewer 2 comments and authors’ response\\
%Reviewer 3 comments and authors’ response
%}

%%%%%%%%%%%%%%%%%%%%%%%%%%%%%%%%%%%%%%%%%%
\end{document}